\documentclass[conference]{IEEEtran}
\IEEEoverridecommandlockouts
\usepackage{cite}
\usepackage{amsmath,amssymb,amsfonts}
\usepackage{algorithmic}
\usepackage{graphicx}
\usepackage{textcomp}
\usepackage{xcolor}

\ifCLASSOPTIONcompsoc
\usepackage[caption=false,font=normalsize,labelfont=sf,textfont=sf]{subfig}
\else
\usepackage[caption=false,font=footnotesize]{subfig}
\fi

\def\BibTeX{{\rm B\kern-.05em{\sc i\kern-.025em b}\kern-.08em
    T\kern-.1667em\lower.7ex\hbox{E}\kern-.125emX}}
\begin{document}

\title{End-to-End Lip Reading in Romanian with Cross-Lingual Domain Adaptation and Lateral Inhibition}

\author{
\IEEEauthorblockN{
Emilian-Claudiu Mănescu\IEEEauthorrefmark{2}\IEEEauthorrefmark{1}, Răzvan-Alexandru Smădu\IEEEauthorrefmark{3}\IEEEauthorrefmark{1}, Andrei-Marius Avram\IEEEauthorrefmark{3}, \\
Dumitru-Clementin Cercel\IEEEauthorrefmark{3}\IEEEauthorrefmark{4}, and Florin Pop\IEEEauthorrefmark{3}\IEEEauthorrefmark{5}
}
\IEEEauthorblockA{
\IEEEauthorrefmark{2}Faculty of Mathematics and Computer Science, University of Bucharest, Romania\\
}
\IEEEauthorblockA{
\IEEEauthorrefmark{3}Faculty of Automatic Control and Computers, University Politehnica of Bucharest, Romania\\
\IEEEauthorrefmark{5}National Institute for Research and Development in Informatics - ICI Bucharest, Romania \\
emilian-claudiu.manescu@s.unibuc.ro, \{razvan.smadu,andrei\_marius.avram\}@stud.acs.upb.ro,\\
\{dumitru.cercel,florin.pop\}@upb.ro
}
}

\maketitle

\let\thefootnote\relax\footnotetext{\IEEEauthorrefmark{1} Equal contribution.}
\footnotetext{\IEEEauthorrefmark{4} Corresponding author.}

\begin{abstract}
Lip reading or visual speech recognition has gained significant attention in recent years, particularly because of hardware development and innovations in computer vision. While considerable progress has been obtained, most models have only been tested on a few large-scale datasets. This work addresses this shortcoming by analyzing several architectures and optimizations on the underrepresented, short-scale Romanian language dataset called Wild LRRo. Most notably, we compare different backend modules, demonstrating the effectiveness of adding ample regularization methods. We obtain state-of-the-art results using our proposed method, namely cross-lingual domain adaptation and unlabeled videos from English and German datasets to help the model learn language-invariant features. Lastly, we assess the performance of adding a layer inspired by the neural inhibition mechanism.
\end{abstract}


\section{Introduction}
\label{sec:intro}

Lip reading is the task of understanding and classifying the spoken words from a video based solely on the visual cues of the speaker's lip movements~\cite{chung2016lip,FERNANDEZLOPEZ201853,son2017lip}. It is a challenging task, mainly due to the discrepancy between the number of phonemes (i.e., the smallest unit of sound in language) and visemes (i.e., the visual equivalent of phonemes), which have a many-to-one mapping. The absence of information from the tongue and throat influences the sound being produced. For example, lip reading systems cannot distinguish between `\textit{ma}', `\textit{pa}', and `\textit{ba}' \cite{chung2016lip}. Lip reading has the same ultimate goal as automatic speech recognition (ASR), namely transcribing speech. However, lip reading is more challenging because the input is a sequence of images that introduces spatial and temporal complexity, hence high-dimensional inputs. Instead, the input is a one-dimensional audio waveform in ASR, thus with lower dimensionality~\cite{Prajwal_2022_CVPR}. 

In the past, another challenge was the computational requirements to train such systems~\cite{Prajwal_2022_CVPR}. Over the last few years, lip reading has gained relevance mainly due to increased available computational resources and its inherent usefulness in real-world applications such as video surveillance systems~\cite{7973348}, assisting hearing-impaired persons~\cite{10.1016/j.eswa.2010.09.119}, and biometric identification~\cite{7156322}. 

Considering the type of input, lip reading datasets are classified into character-level, viseme-level, digit-level, word-level, phrase-level, and sentence-level~\cite{9272286,oghbaie2021advances}. Recent approaches~\cite{chung2016lip,app9081599,huang2022novel,prajwal2022sub} addressed the word-level lip reading task, achieving high accuracy. These models were trained on large-scale datasets such as LRW~\cite{chung2016lip} (English language), CAS-VSR-W1k (previously named LRW-1000)~\cite{yang2019lrw} (Chinese language), and GLips~\cite{schwiebert2022multimodal} (German language). However, these approaches are less effective on smaller datasets; for example, in less common languages such as Romanian. With fewer examples, the models require learning complex and robust features. It naturally raises the question of whether it is possible to use the existing data to get higher accuracy on other small datasets, even if it comes from a different language.

For this reason, we propose a Cross-Lingual Domain Adaptation (CLDA) approach that leverages the cross-lingual knowledge acquired during training to improve the performance on other small training datasets, such as LRRo~\cite{jitaru2020lrro}. Our main architecture utilizes the ResNet-18 backbone~\cite{he2016deep} for feature extraction and a stack of multi-stage temporal convolutional neural networks (MS-TCNs)~\cite{farha2019ms} or Bidirectional Gated Recurrent Units (BiGRUs)~\cite{cho-etal-2014-learning} followed by a lateral inhibition layer~\cite
{DBLP:conf/semeval/Pais22}. We evaluate our approach in multiple configurations and assess the importance of using the cross-lingual setting in the neural architecture. The main contributions of our work are as follows:
\begin{itemize}
    \item We propose a model architecture that obtains state-of-the-art performance on the Wild LRRo dataset by performing extensive experiments with various configurations.
    \item We are the first to integrate the lateral inhibition layer in cross-lingual domain adaptation for lip reading.
    \item We assess the performance of adversarially trained models to learn language-invariant features using unlabeled data from several other languages in both bilingual and multilingual setups.
\end{itemize}

\section{Related Work}
\label{sec:related}

\subsection{Lip Reading}
Before the advent of deep learning techniques, the standard lip reading task was mainly based on hand-engineered features obtained, for example, using Hidden Markov Model (HMM)~\cite{cox2008challenge,dupont2000audio,oghbaie2021advances}. Since the development of deep learning approaches, many works~\cite{koller2015deep,7472852} evaluated the performances of  Convolutional Neural Networks (CNNs)~\cite{kim-2014-convolutional}, Long Short-Term Memory (LSTM)~\cite{hochreiter1997long} networks, and architectures~\cite{chung2016lip,chen2020lipreading,martinez2020lipreading} specifically designed for lip reading. For example, Koller et al.~\cite{koller2015deep} used a CNN model to learn the mouth shape and fed the output to an HMM for temporal alignment. Wand et al.~\cite{7472852} extensively evaluated the performance of LSTM networks and obtained significant improvements over support vector machines~\cite{hearst1998support}.

From the VGG-M network~\cite{chatfield2014return}, Chung et al.~\cite{chung2016lip} proposed four novel architectures by incorporating early fusion and multiple towers combined with 3D or 2D convolutional frontend modules. This approach was evaluated on a large-scale dataset, improving existing performances. Stafylakis et al.~\cite{stafylakis2017combining} proposed a network containing 3D convolutions as a spatio-temporal frontend, followed by a 34-layer ResNet and a bidirectional LSTM. Their approach improved the existing state-of-the-art accuracy by almost 7\%. Moreover, Yang et al.~\cite{yang2019lrw} compared frontend architectures when introducing the LRW-1000 dataset and showed that combining 3D and 2D convolutions achieves the best performance. Martinez et al.~\cite{martinez2020lipreading} improved the ResNet-BiGRU network by substituting BiGRU layers with Temporal Convolutional Networks (TCN)~\cite{bai2018empirical}.

Feng et al.~\cite{feng2020learn} introduced a comprehensive study of the most efficient lip-reading training strategies. Most approaches used ResNet-18 for the frontend and recurrent neural networks (RNNs) or MS-TCN for the backend. Ma et al.~\cite{ma2022training} investigated augmentation techniques and evaluated self-distillation~\cite{zhang2019your} with TCN and BiGRU-based architectures. Prajwal et al.~\cite{Prajwal_2022_CVPR} proposed the attention mechanism for the lip reading task, achieving state-of-the-art results on large-scale datasets.

\subsection{Cross-Lingual Domain Adaptation}
Domain adaptation and cross-lingual learning are currently underrepresented in the lip-reading field. Jitaru et al.~\cite{jitaru2021toward} proposed a transfer learning approach by creating a multilingual dataset containing Romanian, English, and Chinese samples to verify if the models could learn language-independent representations. There have been attempts to use the language discriminator outside the visual speech recognition task. For example, Abdelwahab et al.~\cite{abdelwahab2018domain} used unlabeled data from a related dataset to improve acoustic emotion recognition by employing domain adversarial neural networks. Takashima et al.~\cite{takashima2021unsupervised} enhanced the models’ performance on unknown classes by combining unsupervised domain adaptation and knowledge distillation~\cite{hinton2015distilling} in a cross-modal setting. Since audio data provides more information about the words, a teacher model trained on the audio transfers knowledge to the lip-reading student. 
Additionally, lip movement is aligned to the sub-word-level representation of the intermediate layers for the speech input signal to generalize to unknown classes.

Kim et al.~\cite{kim2022speaker} proposed a lip-reading framework based on speaker adaptation. Their approach introduced user-dependent padding that does not require modifying existing pre-trained models. Their framework can be combined with any unsupervised domain adaptation algorithm, managing lower word error rates and higher accuracies than other methods. Domain adaptation was also performed between real and synthetically generated data on a variational encoder model~\cite{DBLP:conf/bmvc/SenAMNJ21}, improving the top-5 accuracy by 20\%.

\begin{figure*}[!ht]
\centering
\subfloat[The baseline lip reading model with BiGRU backend.]{
    \includegraphics[width=0.99\columnwidth]{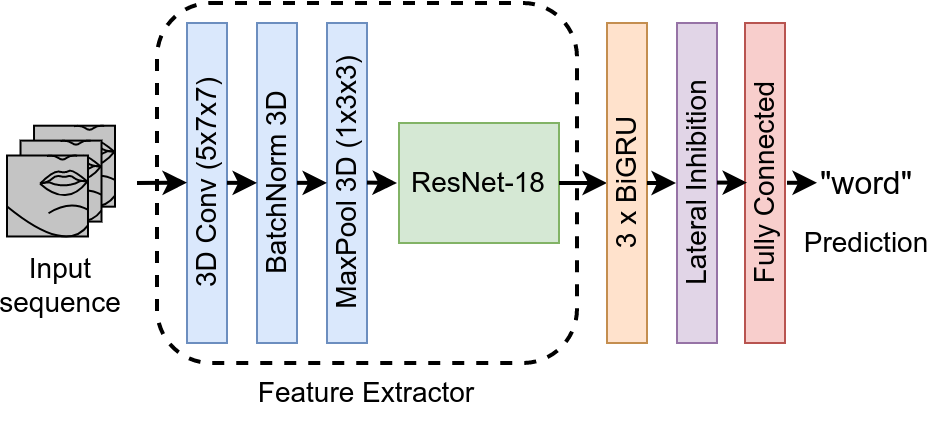}
    \label{fig:std_lr_gru}
}
\subfloat[The baseline lip reading model with MS-TCN backend.]{
    \includegraphics[width=0.99\columnwidth]{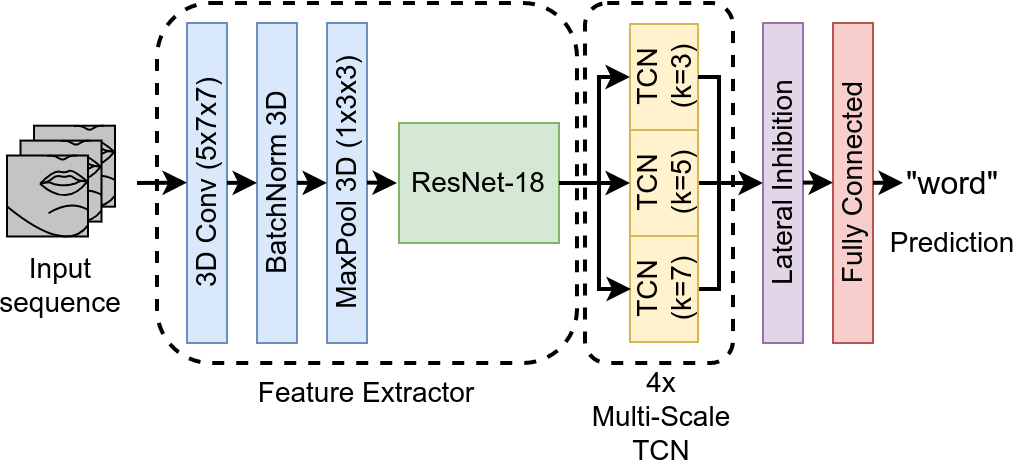}
    \label{fig:std_lr_tcn}
}
\caption{Overview of the baseline lip reading models we used.  ReLU activation functions follow BatchNorm3D layers. The ResNet-18 block ends with a global average pooling layer and a batch norm. The MS-TCN backend consists of four  Multi-Scale TCN layers, while the BiGRU backend consists of three BiGRU layers. Optionally, both models use lateral inhibition before the output layer.}
\end{figure*}

\begin{figure}[!ht]
\centering
\includegraphics[width=\columnwidth]{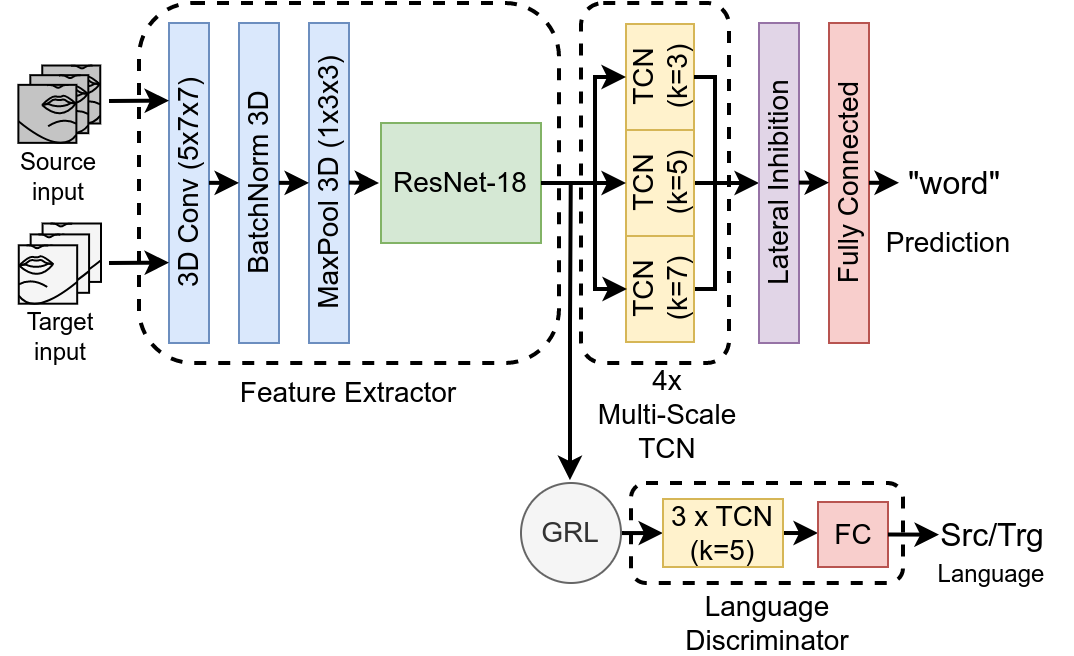}
\caption{Overview of our model trained using CLDA. The input source and target are from different languages.}
\label{fig:da_lr}
\end{figure}

\section{Proposed Method}
\label{sec:method}

\subsection{Lateral Inhibition Layer}

The lateral inhibition (LI) layer~\cite{DBLP:conf/semeval/Pais22} takes inspiration from the biological process of neural inhibition found in the human brain, where the exciting neurons reduce the activity of their neighboring neurons~\cite{Cohen2011}. In the visual cortex, it improves the perception in difficult scenarios such as low-lighting conditions~\cite{avram2022species}. Intuitively, we envisage that the LI layer should help the neural network to focus better on the actual lip reading task while possibly removing unwanted noise. This layer has been successfully applied in several named entity recognition tasks~\cite{avram-etal-2022-racai,avram2022species}, in general, improving performances.

Like a linear layer, the LI layer is defined by the trainable weights $W \in \mathbb{R}^{N \times N}$ and a bias term $b \in \mathbb{R}^{1 \times N}$, where $N$ represents the number of features.
The layer models the input activations $x \in \mathbb{R}^{1 \times N}$ as follows:
\begin{equation}
    \label{eq:li_def}
    LI(x) = x \cdot Diag(\Theta(x \cdot ZeroDiag(W) + b))
\end{equation}
where $Diag(\cdot)$ returns an $N \times N$ matrix with the diagonal equal to the input vector, $ZeroDiag(W)$ zeros the values found on the diagonal of the weight matrix $W$, and $\Theta$ is the Heaviside function described by Equation \ref{eq:heaviside} whose role is to determine which values are allowed to pass to the next layer.

\begin{equation}
\label{eq:heaviside}
\Theta(x) = \left\{
\begin{matrix}
	1, x > 0  \\
	0, x \leq 0
\end{matrix}
\right.
\end{equation}

Because the Heaviside function is non-differentiable in the backward pass, it was replaced with the parameterized sigmoid function~\cite{wunderlich2021event} defined as:
\begin{equation}
    \label{eq:sigmoid}
    \sigma_k(x) = \frac{1}{1+e^{-kx}}
\end{equation}
where $k$ is the scaling parameter. This function is fully-differentiable, the derivative of $\sigma$ depending on $k$ being:
\begin{equation}
    \label{eq:sigmoid_deriv}
    \sigma_k'(x) = k\sigma(x)\sigma(-x)
\end{equation}

This technique is known as surrogate gradient learning~\cite{8891809}, and it enables the use of a known approximate derivative (i.e., Equation \ref{eq:sigmoid_deriv}) in the backward pass as a replacement for the non-differentiable function in the forward pass.

\subsection{Baselines}
For the end-to-end lip reading baseline models, we draw inspiration from the work of Feng et al.~\cite{feng2020learn} and Martinez et al.~\cite{martinez2020lipreading}, whose architectures have achieved state-of-the-art performances. These models are composed of a frontend and a backend architecture. The depiction of the baselines can be seen in Figs.~\ref{fig:std_lr_gru} and \ref{fig:std_lr_tcn}.

\textbf{Frontend Architecture.} The baseline models share the same frontend architecture, consisting of a 3D convolutional block followed by ResNet-18. The convolutional block is a stack of 3D spatio-temporal convolution, BatchNorm 3D, ReLU activation, and max pooling layer. The temporal dimension of the input (i.e., the number of frames of the video) is preserved during processing. Afterward, we pass the obtained ResNet features through a global average pooling layer over the height and width dimensions, which results in a one-dimensional feature tensor for each video frame. Then, we apply batch normalization, resulting in the feature representation of the input. The frontend module only learns independent feature maps without considering the videos' sequential nature.

\textbf{Backend Architectures.} To model the temporal correlation of the frames, we compare two different backend architectures: a three-layer BiGRU network and an MS-TCN system. MS-TCNs add a layer of complexity on top of the typical temporal convolution by simultaneously incorporating several kernels for each layer, thereby combining short- and long-term information while obtaining more robust feature encodings. Using TCNs as an alternative to BiGRUs is justified by the gradient stability, low memory consumption during training, and inherent parallelism, as opposed to RNNs which perform sequential computation. We use the same MS-TCN setup as Martinez et al.~\cite{martinez2020lipreading}, consisting of four layers with three kernels per layer of sizes 3, 5, and 7. Either backend is followed by an optional LI layer and a fully connected output layer for prediction.

\subsection{Cross-Lingual Domain Adaptation}

We incorporate the adversarial domain framework into the baselines to adapt the model to learn language-invariant features and increase robustness. More specifically, the feature representation generated by the last layer of the feature extractor is passed to a language discriminator besides the label predictor. It comprises a stack of three TCN layers with a kernel size of 5, followed by a fully connected layer. The language discriminator receives as input feature representations for both source and target domains (i.e., from two distinct languages). The intuition is to align common representations of similar words of different languages in the feature space while forcing the discriminator not to be able to distinguish between those representations. 

We insert a gradient reversal layer (GRL) before the language discriminator to achieve the minimax objective. The GRL layer acts as the identity function during the forward pass and reverses the sign of the gradients scaled by a hyperparameter $\lambda$ during the backward pass. Therefore, the final loss is defined as follows:
\begin{equation}
L_{DA}(\theta_{f}, \theta_{y}, \theta_{d}) = L_{y}(\theta_{f}, \theta_{y}) - \lambda L_{d}(\theta_{f}, \theta_{d})
\end{equation}
where $L_{y}$ is the lip reading predictor's loss, $L_{d}$ is the language classifier's loss, weighted by the parameter $\lambda$. Fig.~\ref{fig:da_lr} presents the architecture for the lip reading model using cross-lingual domain adaptation (CLDA).

\begin{table*}[!t]
\center
\caption{Baseline results on the Wild LRRo dataset, comparing the influence of DropBlock. Bold indicates the best scores.}
\resizebox{0.7\textwidth}{!}{
\begin{tabular}{| l | c | c | c | c | c | c | c |}
\hline
\textbf{Model} & \textbf{DropBlock} &  \textbf{Acc@1} & \textbf{Acc@5} & \textbf{P(\%)} & \textbf{R(\%)} & \textbf{F1(\%)} \\\hline
\hline
ResNet \& MS-TCN \& LI   & Yes & 45.8$\pm$6.5 & 70.3$\pm$6.0 & 44.1$\pm$4.1 & 41.3$\pm$4.3 & 40.6$\pm$4.1 \\\hline
ResNet \& MS-TCN         & Yes & 43.5$\pm$4.2 & 71.2$\pm$4.7 & 44.8$\pm$3.2 & 42.3$\pm$3.5 & 41.6$\pm$3.0 \\\hline
ResNet \& MS-TCN         & No  & 44.4$\pm$5.5 & 69.8$\pm$4.2 & 41.6$\pm$5.4 & 41.7$\pm$4.2 & 39.8$\pm$4.4 \\\hline
\hline
ResNet \& BiGRU \& LI    & Yes & \textbf{49.1$\pm$6.9} & \textbf{72.2$\pm$6.9} & \textbf{47.8$\pm$4.9} & \textbf{45.1$\pm$2.9} & \textbf{43.8$\pm$3.3} \\\hline
ResNet \& BiGRU          & Yes & 41.6$\pm$2.2 & 67.6$\pm$4.4 & 43.5$\pm$6.3 & 41.2$\pm$2.4 & 39.3$\pm$2.5 \\\hline
ResNet \& BiGRU          & No  & 42.8$\pm$3.5 & 64.7$\pm$5.9 & 44.8$\pm$5.2 & 40.6$\pm$1.3 & 39.5$\pm$2.6 \\\hline
\hline
Inception-V4 \cite{jitaru2020lrro} & No & 33.0 & 62.0 & - & - & - \\
\hline
\end{tabular}}
\label{tab:lipreadstandard}
\end{table*}

\begin{table}[!htp]
\center
\small
\caption{Results on the Wild LRRo dataset for the baseline and CLDA models. Bold indicates the best scores. }
\resizebox{\columnwidth}{!}{
\begin{tabular}{| l | c | c | c |}
\hline
\textbf{Model} & \textbf{Languages} & \textbf{Acc@1} & \textbf{Acc@5} \\\hline
\hline
ResNet \& MS-TCN          & ro & 43.5$\pm$4.2 & 71.2$\pm$4.7 \\\hline
ResNet \& MS-TCN \& LI    & ro & 45.8$\pm$6.5 & 70.3$\pm$6.0 \\\hline
ResNet \& BiGRU           & ro & 41.6$\pm$2.2 & 67.6$\pm$4.4 \\\hline
ResNet \& BiGRU \& LI     & ro & \textbf{49.1$\pm$6.9} & \textbf{72.2$\pm$6.9} \\\hline
\hline
ResNet \& MS-TCN          & ro \& en & 47.5$\pm$1.1 & 72.7$\pm$0.7 \\\hline
ResNet \& MS-TCN \& LI    & ro \& en & 48.9$\pm$2.1 & 73.2$\pm$3.5 \\\hline
ResNet \& BiGRU           & ro \& en & 47.9$\pm$4.2 & \textbf{73.9$\pm$4.7} \\\hline
ResNet \& BiGRU \& LI     & ro \& en & \textbf{49.0$\pm$6.5} & 72.3$\pm$3.3 \\\hline
\hline
ResNet \& MS-TCN          & ro \& de & 47.3$\pm$4.7 & 72.9$\pm$2.8 \\\hline
ResNet \& MS-TCN \& LI    & ro \& de & \textbf{51.6$\pm$4.3} & \textbf{74.6$\pm$3.0} \\\hline
ResNet \& BiGRU           & ro \& de & 44.0$\pm$6.5 & 70.5$\pm$9.5 \\\hline
ResNet \& BiGRU \& LI    & ro \& de & 48.4$\pm$2.8 & 73.6$\pm$3.3 \\\hline
\hline
ResNet \& MS-TCN          & ro \& en \& de & 48.6$\pm$6.0 & 73.1$\pm$5.7 \\\hline
ResNet \& MS-TCN \& LI    & ro \& en \& de & 45.8$\pm$4.1 & 71.4$\pm$2.8 \\\hline
ResNet \& BiGRU           & ro \& en \& de & 46.6$\pm$5.6 & 68.6$\pm$4.7 \\\hline
ResNet \& BiGRU \& LI     & ro \& en \& de & \textbf{48.8$\pm$4.3} & \textbf{76.0$\pm$4.6} \\\hline
\hline
Inception-V4 \cite{jitaru2020lrro}  & - & 33.0 & 62.0 \\
\hline
\end{tabular}}
\label{tab:languages_comparison}
\end{table}

\subsection{Optimization Strategies}
Due to the limited size of the datasets to evaluate our method, we employ a variety of optimizations intended to reduce overfitting:
\begin{itemize}
  \item \textit{Data augmentation} via random horizontal flip and mixup~\cite{zhang2017mixup}.
  \item \textit{Label smoothing}~\cite{muller2019does}. Models are prevented from making overconfident predictions through soft class probabilities rather than hard labels (0 or 1).
  \item \textit{Squeeze-and-Excitation blocks}~\cite{hu2018squeeze}. Each input channel is assigned a different weight, which will scale the filter maps of each convolutional kernel accordingly.
  \item \textit{Cosine Learning Rate scheduler}. It ensures that both low and high values for the learning rate are used, namely the high values to avoid getting stuck in the local minimum and the low values to ensure the model slowly finds the optimal minimum.
  \item \textit{DropBlock regularization}~\cite{ghiasi2018dropblock}. It drops activations in a continuous region to ensure spatial information does not leak into the following layers.
\end{itemize}

\section{Experiments}
\label{sec:experiments}

\subsection{Datasets}
To evaluate our approach, we utilize word-level datasets from three languages: Romanian with the LRRo dataset~\cite{jitaru2020lrro}, English with the LRW dataset~\cite{chung2016lip}, and German with the GLip dataset~\cite{schwiebert2022multimodal}.

\textbf{LRRo.} The LRRo dataset is available in two versions: Lab LRRo and Wild LRRo. Lab LRRo has video frames recorded in a controlled environment with similar recording angles and lighting conditions. The Wild LRRo dataset contains images of 21 words spoken by 35 persons from publicly available sources, such as news and TV shows, captured in various conditions. In total, the Wild LRRo dataset consists of 1,087 samples split into train (with 846 examples), validation (with 120 examples), and test (with 121 examples) sets. Currently, LRRo is the only Romanian dataset of its kind. 

\textbf{LRW.} The Lip Reading in the Wild (LRW) dataset contains English words spoken on television shows from the BBC\footnote{British Broadcasting Corporation} channel. It contains 500 different words, each spoken by over 1,000 persons. Each video is segmented into 29 frames (about 1.16 seconds) per word. The length of the words varies between 5 and 12 letters. The dataset is split as follows: over 800 samples per word are assigned in the training set, and 50 samples per word are used in the validation and test.

\textbf{GLips.} The German Lips (GLips) dataset contains more than 250,000 videos from 100 speakers, recorded in various conditions from recorded parliamentary sessions. Like the LRW dataset, GLips has 500 different words, sized between 4 and 18 letters, spoken in a 1.16 seconds window frame. Each word has 500 instances, resulting in 250,000 examples in total. The dataset is split into train (200,000 samples), validation (50,000 samples), and test (50,000 samples) sets.

We evaluate our models on the Wild LRRo dataset. For the LRW and GLips datasets, we randomly extracted a subset with an equal number of samples to ensure a balanced distribution. This process resulted in 848 examples for training, 120 for validation, and 121 for testing. We use the LRW and GLips datasets, respectively, for the domain adaptation experiments as the second dataset.

\begin{figure*}[!ht]
\centering
\subfloat[ResNet \& BiGRU \& LI trained on the Romanian language.]{
    \includegraphics[width=0.99\textwidth]{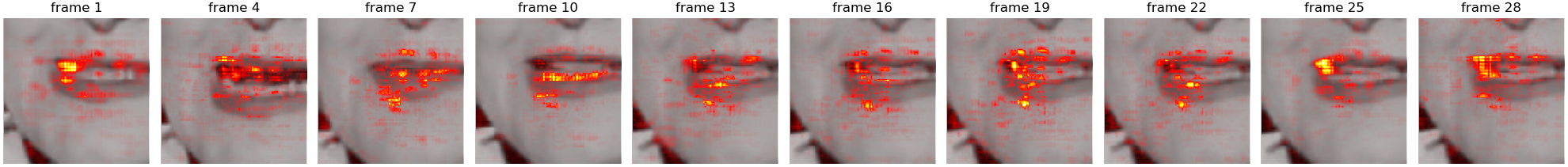}
}
\hfil
\subfloat[ResNet \& BiGRU \& LI trained on the Romanian and English languages, in domain adaptation setting.]{
    \includegraphics[width=0.99\textwidth]{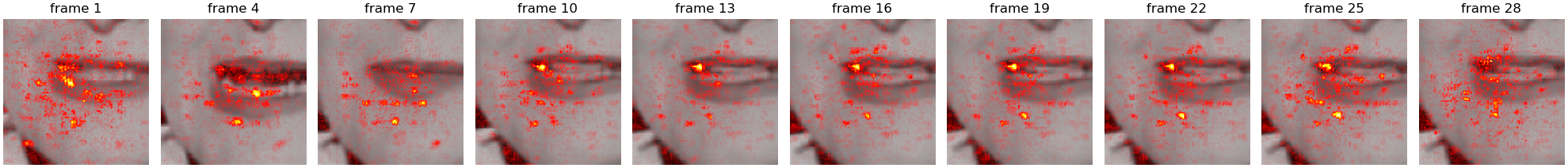}
}
\hfil
\subfloat[ResNet \& MS-TCN \& LI trained on the Romanian and German languages, in domain adaptation setting.]{
    \includegraphics[width=0.99\textwidth]{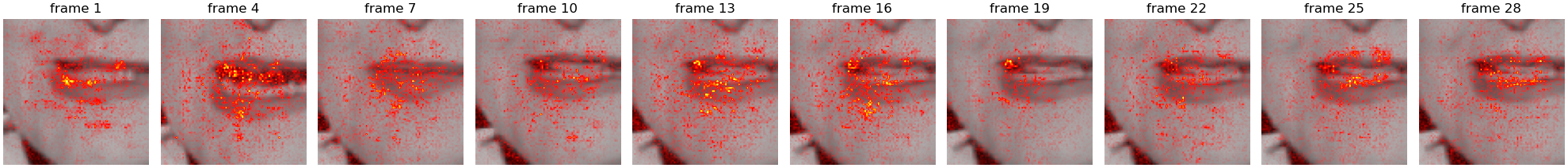}
}
\hfil
\subfloat[ResNet \& BiGRU \& LI trained on the Romanian, English, and German languages, in domain adaptation setting.]{
    \includegraphics[width=0.99\textwidth]{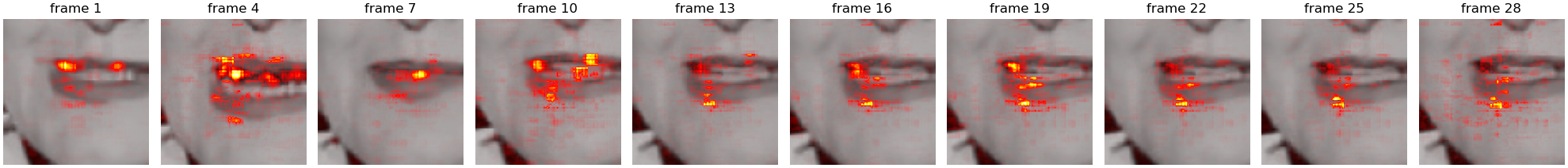}
}
\caption{\label{fig:saliency_maps} Saliency maps generated by the best models on the test example for the word ``inseamna'' (Engl. ``it means''). The brighter color indicates a higher neuron response in that region contributing to the prediction label. Best viewed in color.}
\end{figure*}

\begin{figure}[!ht]
    \centering
    \includegraphics[width=\columnwidth]{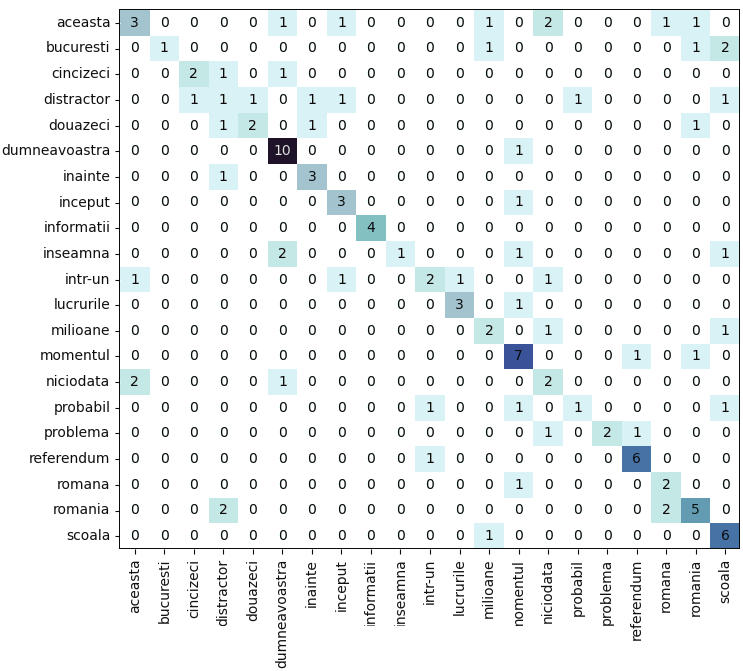}
    \caption{Confusion matrix for ResNet \& MS-TCN in multilingual setting.}
    \label{fig:cm}
\end{figure}

\subsection{Experimental Setup}

We train the model using the Adam optimizer~\cite{adam2015diederik} with a weight decay of $1e-4$ and an initial learning rate of $3e-4$ for the cosine learning rate scheduler. The batch size is kept between 8 and 32, depending on the available GPU memory. The dropout probability is set to 12\%, and the BiGRU hidden size is 512. We train the models for at most 80 epochs. Experiments were performed on Nvidia P100 and T4 GPUs, taking on average between one and two hours to train per experiment. We report top-1 accuracy (Acc@1), top-5 accuracy (Acc@5), precision (P), recall (R), and weighted F1-score (F1) reported over five runs as mean and standard deviation values. For discussions, we select the best models with the highest top-1 accuracy on the validation set.

\subsection{Results}
\subsubsection{Baselines}
The results obtained for the baseline lip reading models are presented in Table~\ref{tab:lipreadstandard}. The model using the DropBlock regularization, BiGRU, and lateral inhibition obtained the best performance, surpassing the model without DropBlock by 7-8\% in both top-1 and top-5 accuracy. We also note that the MS-TCN backend outperforms the BiGRU backend, especially in top-5 accuracy, when lateral inhibition is not employed, albeit, with almost double model size and 35 times more parameters to train (i.e., 11.3M vs. 379M).

Due to the small size of the Wild LRRo dataset and the complex nature of the task, we expected regularization to play an important part in increasing the models' performances. We mainly evaluate the impact of adding the DropBlock regularization since, to the best of our knowledge, it had never been used before on the lip reading task.
On the other hand, lateral inhibition improves models' performances in the baseline experiments. All ResNet-based models outperform the Inception-V4 results of Jitaru et al.~\cite{jitaru2020lrro}.

\subsubsection{CLDA and Lateral Inhibition}
Experimental results on the CLDA task are summarized in Table~\ref{tab:languages_comparison}. In these experiments, we evaluate with and without including the lateral inhibition layer. We observe that the models can learn language-independent features using unlabeled data from relatively unrelated languages. We believe that the 1-2\% performance difference between the German GLips and English (which come from related Germanic languages) LRW datasets may be due to the manual sampling of the two datasets since the words used in the reduced datasets obtained may use different sets of visemes. The German language employed in ResNet \& MS-TCN achieves higher scores, while the English language employed in the ResNet \& BiGRU architectures performs better. In the multilingual setup, we merged the two support datasets (i.e., GLips and LRW) during training, obtaining 48\% top-1 accuracy and 76\% top-5 accuracy.

We establish new state-of-the-art performance on the Wild LRRo dataset. We see a higher score on the MS-TCN backend when we do not include the LI layer and higher results on the BiGRU backed when we include the LI layer. Employing domain adaptation, we achieve higher overall performances, despite the language. Additionally, including a second language, the model improves the top-5 accuracy by 1.4\%, while the top-1 accuracy is affected.

\section{Discussions and Limitations}
\label{sec:discussions}

The most performance gains were obtained when we incorporated cross-lingual knowledge into the model. The model predicts better the most frequent classes from the dataset, such as ``dumneavoastră'' (a more formal way of saying ``you'' in Romanian), ``momentul'' (Engl. ``the moment''), and ``școală'' (Engl. ``school''). These results can also be seen in the confusion matrix presented in Fig.~\ref{fig:cm}. However, we observe that the models struggle to detect words that share some similar visemes, such as ``aceasta'' (Engl. ``this'') and ``niciodată'' (Engl. ``never'').

We also analyze which parts of the image influence the most when generating the prediction labels by visualizing the salient features~\cite{simonyan2014deep} on the test examples. The results for the best-performing model from each set of experiments are presented in Fig.~\ref{fig:saliency_maps}. We notice that the baseline and the cross-lingual domain adaptation models focus more on the lips. When we employ the domain adaptation setting with either English or German, we notice that the model considers areas from the whole face, with high response around darker spots around the mouth. This result is also generally consistent with the score: best models learn to look for features around the mouth.

The limited size of the datasets we used, especially for the subsets for German and English languages, may hinder the performance. One of our goals was to use limited resources to train high-performing models. The models used in this manuscript can be trained using freely available resources such as Google Colaboratory. One future direction is enabling more data for these languages during the training procedure, as models trained on these large-scale datasets achieve over 60-70\% top-1 accuracy (see Chung et al.~\cite{chung2016lip} and Schwiebert et al.~\cite{schwiebert2022multimodal}). In this regard, the findings of Schwiebert et al.~\cite{schwiebert2022multimodal} regarding improved performances while applying transfer learning align with ours.

\section{Conclusion}
\label{sec:conclusion}

This paper provides a comprehensive overview of the performances of different lip reading neural network architectures adapted to the small-scale Romanian-language Wild LRRo dataset. We compare two widely used backend modules for lip reading, assess the impact of adding lateral inhibition, and show that using DropBlock regularization leads to a remarkable increase in accuracy. We also propose an adversarially-trained architecture that obtains state-of-the-art results by using unlabeled videos from other languages, helping the model to language-invariant features. For future work, we aim to investigate how lateral inhibition affects the future space.

\section*{Acknowledgments}
This research has been funded by the University Politehnica of Bucharest through the PubArt program.

\bibliographystyle{IEEEtran}
\bibliography{mybib}

\end{document}